\documentclass{vgtc}                          

\vgtcinsertpkg

\graphicspath{{figures/}{pictures/}{images/}{./}} 

\usepackage{times}                     

\usepackage{tabu}                      
\usepackage{booktabs}                  
\usepackage{lipsum}                    
\usepackage{mwe}                       

\usepackage{mathptmx}                  

\usepackage{subfig}
\usepackage[ruled,vlined,algo2e]{algorithm2e}
\usepackage{enumitem}
\usepackage{soul}
\usepackage{xcolor}
\usepackage{graphicx}

\usepackage{url}

\usepackage{breakurl}

\usepackage{tikz}
\newcommand*\circled[1]{\tikz[baseline=(char.base)]{\node[shape=circle,draw,fill=black,text=white,inner sep=0.0pt, minimum size=1em,minimum width=1em,minimum height=1em] (char) {#1};}}

\makeatletter
\renewcommand{\@fnsymbol}[1]{}
\makeatother

\newcommand{\fixme}[1]{{{#1}}}
\renewcommand{\hl}[1]{#1}

\onlineid{3094}

\vgtccategory{Research}

\title{Towards Energy-Efficiency by Navigating the Trilemma of \\ Energy, Latency, and Accuracy}

\author{Boyuan Tian\thanks{}
\and Yihan Pang
\and Muhammad Huzaifa
\and Shenlong Wang
\and Sarita Adve\thanks{e-mail: \{boyuant2, yihanp2, huzaifa2, shenlong, sadve\}@illinois.edu}}
\affiliation{University of Illinois Urbana-Champaign}

\teaser{
  \centering
  \includegraphics[width=\linewidth]{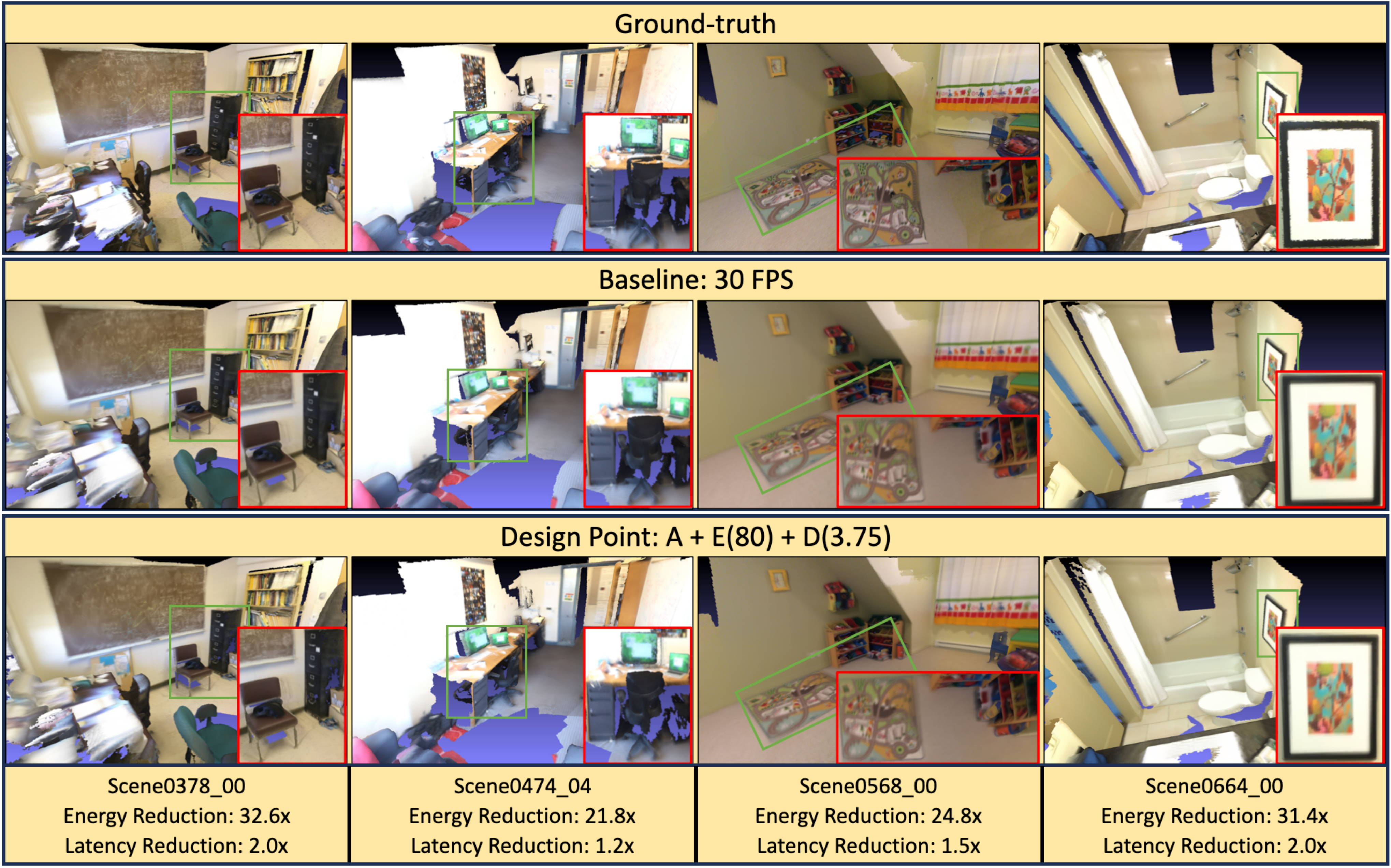}
  \caption{Scene reconstruction achieves significant reductions in energy and latency with negligible accuracy loss. We systematically characterize the trade-off space in energy, accuracy, and latency, enabled by three energy-oriented optimizations, and identify optimal designs under various downstream application imposed constraints. Qualitative results for the ground-truth, the baseline algorithm at 30~FPS, and the chosen design for an application scenario show that meshes from the chosen design are almost indistinguishable from the best achievable quality (Row 2). Disparities between the chosen design and the ground-truth are attributed to the baseline algorithm, as seen in Row 1 and Row 2.
  }
  \label{fig:quality-evaluation}
}

\abstract{
Extended Reality (XR) enables immersive experiences through untethered headsets but suffers from stringent battery and resource constraints. Energy-efficient design is crucial to ensure both longevity and high performance in XR devices. However, latency and accuracy are often prioritized over energy, leading to a gap in achieving energy efficiency. This paper examines scene reconstruction, a key building block for immersive XR experiences, and demonstrates how energy efficiency can be achieved by navigating the trilemma of energy, latency, and accuracy.

We explore three classes of energy-oriented optimizations, covering the algorithm, execution, and data, that reveal a broad design space through configurable parameters. Our resulting 72 designs expose a wide range of latency and energy trade-offs, with a smaller range of accuracy loss. We identify a Pareto-optimal curve and show that the designs on the curve are achievable only through synergistic co-optimization of all three optimization classes and by considering the latency and accuracy needs of downstream scene reconstruction consumers. Our analysis covering various use cases and measurements on an embedded class system shows that, relative to the baseline, our designs offer energy benefits of up to 60$\times$ with potential latency range of 4$\times$ slowdown to 2$\times$ speedup. Detailed exploration of a use case across representative data sequences from ScanNet showed about 25$\times$ energy savings with 1.5$\times$ latency reduction and negligible reconstruction quality loss.
}


\begin{document}
\maketitle

\section{Introduction}
\label{sec:intro}

\hl{Machine} Perception - the capability enabling Extended Reality (XR, the collection of AR/VR/MR) systems to understand and interpret their surroundings - is the cornerstone for high-quality user interactions. It covers a wide range of algorithms, including scene reconstruction, eye tracking, object detection, panoramic segmentation, etc. \hl{Machine} perception algorithms have historically been prioritized for high performance (i.e., \textbf{accuracy}) and low frame processing delay (i.e., \textbf{latency}\footnote{\hl{This is not the end-user's Motion-to-Photon (MTP) latency but rather an algorithm's execution latency, which is often hidden by mechanisms like asynchronous reprojection (timewarp), etc.}}), with insufficient attention paid to \textbf{energy} consumption. Unfortunately, achieving high accuracy and low latency often comes at the cost of substantial energy usage, resulting in a trilemma among energy, latency, and accuracy.

High energy consumption can be detrimental to the immersive experience, especially on mobile XR devices with constrained battery capacity. High energy consumption can quickly deplete the battery, rendering the system inoperable. Algorithms that consume high amounts of energy often need to be offloaded to servers~\cite{dhakal2022slam,jiang2023offloading,lee2023farfetchfusion}, incurring networking delay penalties, or require increased battery capacity with additional weight, neither of which are optimal options. For instance, Apple's Vision Pro~\cite{AppleVisionPro,Smith_2023} is reported to have a battery life of about two hours, which is similar to Meta's Quest Pro~\cite{MetaQuestPro1} but with  $\sim$1.7$\times$ the battery capacity and $\sim$3.4$\times$ the battery weight~\cite{VisionProBatteryCapacity,MetaQuestProBattery}. 

\hl{We study scene reconstruction (spatial mapping) for its desirability in XR and its intense demands on energy and computing resources~\cite{bujanca2022acefusion,dai2017bundlefusion,huzaifa2021illixr}. While historically it was rare to find scene reconstruction deployed on headsets primarily due to resources constraints, it is now supported by high-end XR systems like Vision Pro{~\cite{VisionproSpatialMapping}}, Magic Leap 2{~\cite{MagicleapSpatialMapping}}, and Meta Quest 3{~\cite{QuestSpatialMapping}}. Scene reconstruction is a crucial for enabling immersive visual~\cite{gruber2014efficient,osti2019real} and audio experiences~\cite{kim2022immersive,SpatialAudio,schissler2017acoustic}. Efforts in pursuing energy efficiency for scene reconstruction are therefore likely to benefit future XR systems by enhancing battery life and system usability.}

Scene reconstruction replicates physical objects in virtual space by progressively integrating posed RGB-D frames into a scene representation, such as surfels or voxels. Since the on-device pose tracker is shared by downstream tasks in real XR systems (e.g., Meta Quest~\cite{MetaPoseTracker}), we focus on data integration. Specifically, we examine the Truncated Signed Distance Function (TSDF) Fusion~\cite{curless1996volumetric} in voxel-based scene representation, which merges RGB-D data into a voxel-based truncated distance map. While TSDF Fusion is a well-established algorithm that balances low latency and high accuracy, our study indicates that it is far from being energy-efficient.

We aim for energy efficiency by simultaneously and holistically considering all three dimensions of energy, latency, and accuracy. We first discuss three classes of optimizations with configurable parameters that create a large design space with various trade-offs in the above three dimensions. 

\begin{figure}[t]
    \centering
    \includegraphics[width=\columnwidth]{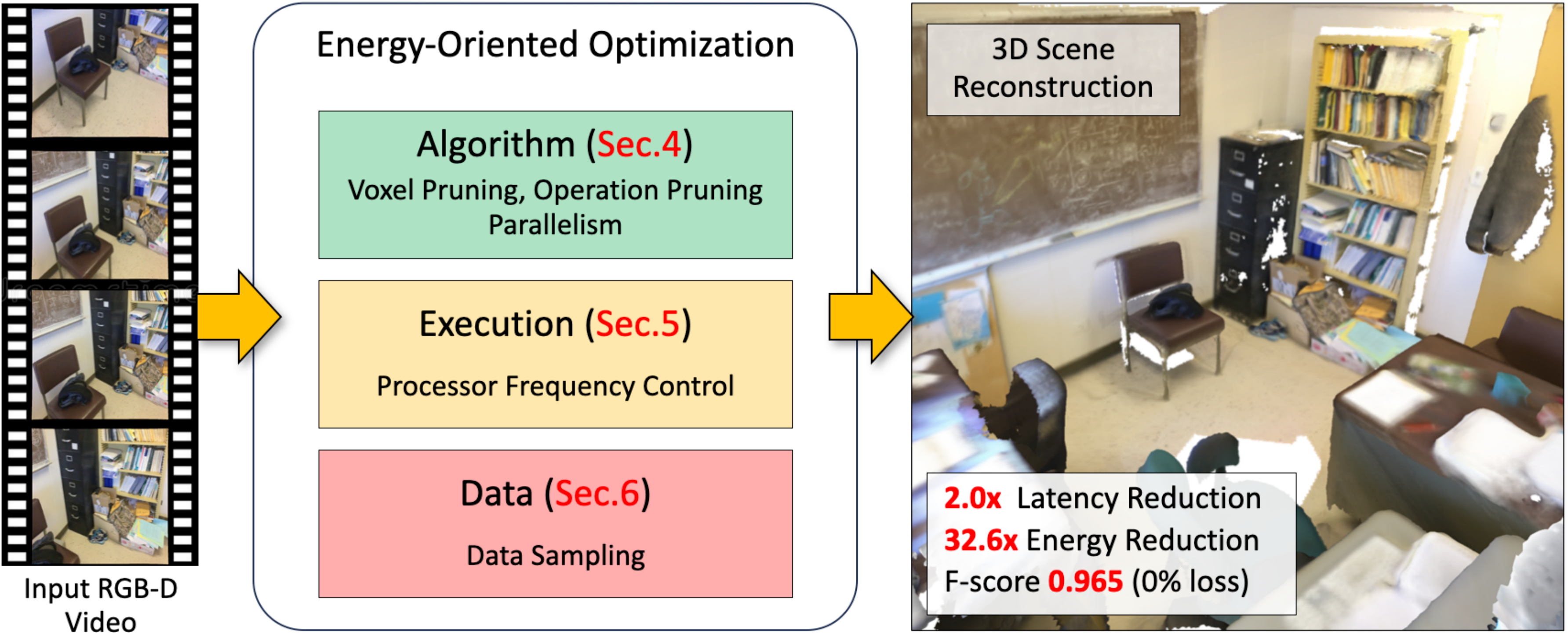}
    \caption{We study three classes of optimizations that create a large design space through configurable parameters, where the designs provide different trade-offs in the dimensions of energy, latency, and accuracy. Co-optimization across all classes leads to significant energy savings within latency and accuracy constraints, illustrated on the right for a case that additionally provides latency benefits with negligible accuracy loss.}
    \label{fig:teaser}
\end{figure}

These optimizations, also summarized in~\cref{fig:teaser} are as follows. 

\begin{itemize}[nolistsep]
    \item Algorithm (A): We identify significant operation redundancies and parallelization overheads in the baseline, which lead to unnecessary energy usage. Our optimizations in this class reduce these inefficiencies, providing lower energy while also reducing latency and without any impact on accuracy.
    \item Execution (E): Processor clock frequency impacts energy and latency without affecting accuracy. While conventional wisdom is to execute as fast as possible, choosing a lower frequency can reduce energy, albeit at the cost of higher latency.
    \item Data (D): Sensory data-based \hl{machine} perception contains rich redundancies across frames. Data sampling can substantially reduce energy but may impact accuracy and latency.
\end{itemize}

We then characterize the energy, latency, and accuracy of 72 designs within the trade-off space on a mobile computing platform using a data sequence from ScanNet. Each design evaluated involves combinations of algorithmic changes, execution frequency, and data sampling rate, with varying configuration parameters. We approach energy efficiency as minimizing energy usage while meeting specified latency and accuracy constraints. \hl{Due to limited low-level hardware accessibility (e.g., for processor frequency and power measurement) on real headsets, we} use the Nvidia Jetson Xavier~\cite{Jetson_xavier_agx} for experimentation. It offers rich computing resources, native power gauge support, and a configurable power budget (TDP~\cite{Jetson_agx_guide}) that can be adjusted to match recent headsets such as the Vision Pro~\cite{VisionProBatteryCapacity} and  Quest Pro~\cite{MetaQuestProBattery}.

We observe three key findings and their implications for achieving energy-efficient design. {\circled{1}} \textbf{Our designs span a wide range of energy and latency trade-offs}. Latency can vary from a 4$\times$ slowdown to a 2$\times$ speedup, with energy savings of up to 60$\times$, relative to our off-the-shelf TSDF Fusion baseline implementation. Designs on the Pareto-optimal curve show a similar range, underscoring the large design space available for optimization and the energy inefficiency of the baseline.
{\circled{2}} \textbf{Co-optimization is essential for achieving designs on the Pareto-optimal curve.} Optimizing processor frequency or data sampling alone is inadequate to reach all designs on the Pareto-optimal curve. Co-optimization can achieve additional energy reductions ranging from 1.32$\times$ to 5.42$\times$. {\circled{3}} \textbf{Applying tailored constraints from cross-component analysis maximizes energy efficiency}. Downstream components have varying latency and accuracy constraints. Identifying and applying appropriate constraints can enable more aggressive energy-efficient designs, potentially yielding additional energy benefits of 1.5$\times$ with relaxed accuracy and 1.66$\times$ with relaxed latency.

We apply our findings and analysis to three downstream use cases—scanning and sharing, spatial audio, and an intermediate scenario—with varying constraints for latency and accuracy. We identify the lowest energy designs for each from our evaluated trade-off space. We find energy benefits of about 16$\times$, 60$\times$, and 27$\times$ respectively for these use cases, relative to our baseline. 

We comprehensively evaluate the sensitivity of our results to input data for the last use case as a representative, studying multiple sequences with various scales, scene types, trajectory patterns, etc. The chosen energy optimal design generalizes well on the ScanNet~\cite{dai2017scannet} indoor SLAM dataset on Jetson AGX Xavier, reducing energy usage by an average of \fixme{24.90$\times$} and respecting the latency and accuracy constraints (frame latency reduces by \fixme{1.45$\times$} and F-score by a negligible (\fixme{0.005}).

While we focus on TSDF Fusion, the insights from navigating the extensive trade-off space to resolve the trilemma between energy, latency, and accuracy are broadly applicable. In summary, this paper makes the following contributions:

\begin{itemize}[nolistsep]
    \item We exploit three classes of optimizations to make trade-offs among energy, latency, and accuracy, exposing a broad trade-off space with designs having distinct priorities.
    \item We characterize the trade-off space and demonstrate that achieving the optimal design requires co-optimizations within an algorithm and across components in a full system.
    \item We show that while optimal design depends on specific use cases, a design identified under certain constraints generalizes well to diverse input data.
\end{itemize}
\section{Related Work}
\label{sec:related}

\textbf{Single-Dimension Enhancements.}
In perception research, significant progress has been made by focusing on improving a single dimension, often without directly considering the impact on other dimensions.
For instance, many studies focus on enhancing accuracy through developing advanced algorithms and neural models, improving object detection precision and recall~\cite{bansal2018zero,bochkovskiy2020yolov4,carion2020end,liang2018deep,wang2015holistic,yin2021center,zohar2023prob}, advancing scene understanding~\cite{chen2020scanrefer,cheng2022masked,graham20183d,huang2018recurrent,kirillov2019panoptic,thomas2019kpconv}, and enhancing 3D scene reconstruction quality~\cite{kerbl20233d,mildenhall2021nerf,muller2022instant, xu2019tigris}. On the other hand, reducing perception latency is crucial for real-time applications like autonomous driving and augmented reality. Recent research includes techniques to reduce perception latency by decreasing computational complexity through methods such as quantization~\cite{chen2023sparsevit,haq}, network pruning~\cite{han2015deep,lin2017runtime}, and more efficient algorithms and neural models~\cite{fan2022embracing,feng2020mesorasi,lin2019tsm}. Finally, efforts to reduce energy consumption include utilizing specialized hardware~\cite{fu2023gen,gan2021eudoxus,lee2023neurex,li2023instant,mubarik2023hardware} in machine perception tasks.

Despite these achievements, focusing solely on single-dimension enhancements can lead to improvements in one area while overlooking trade-offs in others. For example, increasing accuracy may come at the cost of higher computational demands or increased energy consumption. This paper demonstrates the need to co-optimize simultaneously for all dimensions to achieve an optimal design.

\textbf{Multi-Dimensional Trade-off Exploration.} Past research has also explored multi-dimensional optimization. Most related, there is extensive work on the accuracy-latency trade-off for advancing low-power vision~\cite{hu2019banner,huzaifa2023adaptivefusion,likamwa2013energy}, model compression~\cite{han2016eie,he2018amc,li2020gan,ren2023ugc}, and efficient neural networks~\cite{lin2022device,yan2023plenvdb,zhu2023pockengine} for resource-constrained devices. Recent vision competitions~\cite{LPIRC,LPCV} also focus on low-power solutions, where participants are tasked with creating solutions that excel in accuracy and execution time on specific low-power devices. However, energy — defined as the product of power and time — is typically not a primary evaluation criterion in these competitions. As long as the models can run on the designated device, the evaluation centers on which solution provides better accuracy and lower execution time, often overlooking the potential for more energy efficient designs. This paper considers a more holistic and simultaneous exploration of all three dimensions of energy, latency, and accuracy, to identify substantially more energy-efficient designs, with scene reconstruction as our specific target for increased efficiency.
\section{Background}
\label{sec:background}

{\bf TSDF Fusion:} Voxel-based scene reconstruction algorithms represent the scene using voxels, a primitive occupying a cubic region in 3-d space and holding the distance to the nearest surface as Truncated Signed Distance Function (TSDF) values~\cite{zollhofer2018state}. TSDF Fusion is the dominant stage in the algorithm. It merges depth observations from various frames into a unified TSDF value for the same voxel, using the equation:
\begin{equation}
\label{eq:tsdf}
    T_{n} = \frac{W_{n-1} T_{n-1} + w_{n}d_{n}}{W_{n-1} + w_{n}}, \quad
    W_{n} = W_{n-1} + w_{n}
\end{equation}
where $T_n$ and $W_n$ are the TSDF and weight values stored in the voxel after fusing frame $n$, and $w_n$ is the weight used to fuse depth measurement $d_n$.

Adjacent voxels are often grouped into cubic volumes to exploit data locality and reduce the overhead of manipulating millions of voxels in a hash table or an Octree. For instance, Open3D~\cite{zhou2018open3d} and InfiniTAM~\cite{prisacariu2017infinitam} group 4096 ($16^3$) and 512 ($8^3$) voxels, respectively.

TSDF Fusion first checks if a volume is visible from the current camera viewport. 
All voxels in a visible volume are then exhaustively accessed to project onto the depth image plane and assess their eligibility for TSDF calculation. A voxel proceeds to the TSDF calculation phase only if it meets four conditions:
\begin{itemize}[nolistsep]
    \item The voxel locates in front of the camera plane.
    \item The voxel is close to a surface (within the truncation band).
    \item The projected location falls within the image boundaries.
    \item The projected location has a valid depth measurement.
\end{itemize}
Failing to satisfy any of these conditions results in early termination of processing of that voxel.

{\bf Baseline algorithm:} For our baseline algorithm and implementation, we use Open3D v0.16.0, a widely-adopted library developed by Intel known for its low latency and high accuracy. We focus on voxel-based scene reconstruction algorithms, primarily due to their efficient implementation and suitability for mobile devices. It is the de facto choice in both classical~\cite{park2017colored,zhou2016fast} and neural-based~\cite{khoury2017learning} algorithms. We use 10 mm voxels and a truncation band of 10 cm.

{\bf Metrics:} For latency, we report the average time taken for TSDF Fusion per frame in the data sequence (ScanNet sequences in our case). For energy, we report the energy for the entire sequence. Since some of our techniques reduce the number of frames processed, the total processing time for the entire sequence depends on both the number of frames processed and the average latency per frame. In streaming applications, the latency per frame is often more important (e.g., for scheduling) than the entire sequence time. For energy, however, the energy of the entire run determines battery life. We therefore report latency as the average per-frame and report energy as the total for the sequence. Finally, for accuracy, we use the F-score~\cite{Guo_2022_CVPR,Ju_2023_ICCV} to quantify the quality of the final reconstructed mesh against the ground-truth meshes from the dataset.
\section{Algorithm Driven Energy Optimization}
\label{sec:algorithm}

The most natural way to reduce energy consumption is to reduce the work done to perform the specific machine perception task. This typically results in reduced latency and often without loss of accuracy, with a win-win situation in all dimensions. Work reduction can be achieved by reducing the complexity of the algorithm and also by addressing sources of computation waste. We consider three examples below for TSDF Fusion. 

\subsection{Voxel Pruning}
Computation is wasted when it does not yield useful results; e.g., multiple checks for voxels that do not proceed to TSDF calculation. We first characterize the eligibility of voxels for TSDF Fusion, and then illustrate our design for early pruning of ineligible voxels.

\textbf{Characterization.}
We record the status of each voxel during processing in the baseline algorithm. A voxel may undergo processing multiple times within a frame and across a sequence. Specifically, the processing of each voxel results in either a fused voxel (V. Fused) or a non-fused voxel for four reasons: (1) the voxel is behind the camera plane (V. Behind Cam. Plane); (2) the voxel is outside the truncation band (V. Out of Trunc. Band); (3) the projected location is outside the image scope (P. Out of Image Scope); and (4) the projected location has invalid depth value (P. Invalid Depth Value).

\begin{figure}[t!]
  \begin{minipage}[t]{0.40\columnwidth}
    \centering
    \includegraphics[width=\columnwidth]{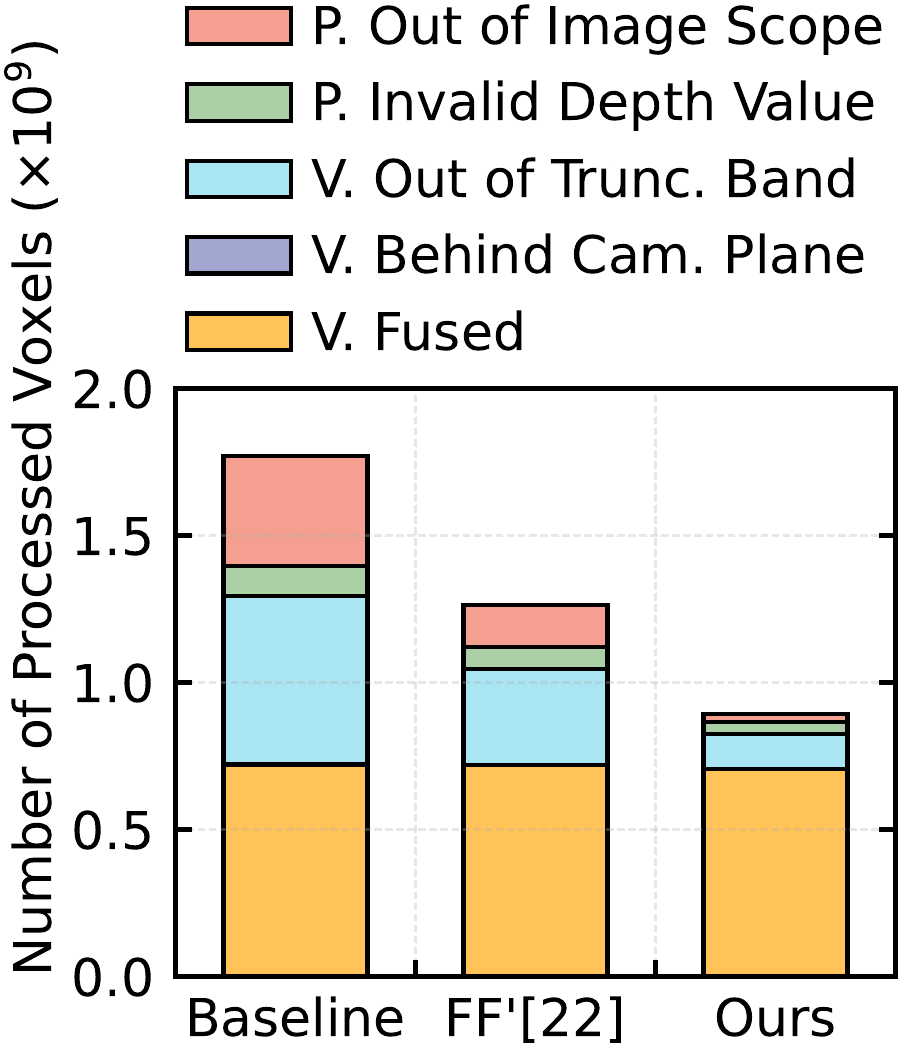}
    \caption{The breakdown of processed voxel status shows that only about 40\% of voxels are fused (V. Fused). Our design filters out non-contributing voxels while preserving the critical ones nearly intact.}
    \label{fig:fusion-char}
  \end{minipage}
  \hfill
  \begin{minipage}[t]{0.57\columnwidth}
    \centering
    \includegraphics[width=\columnwidth]{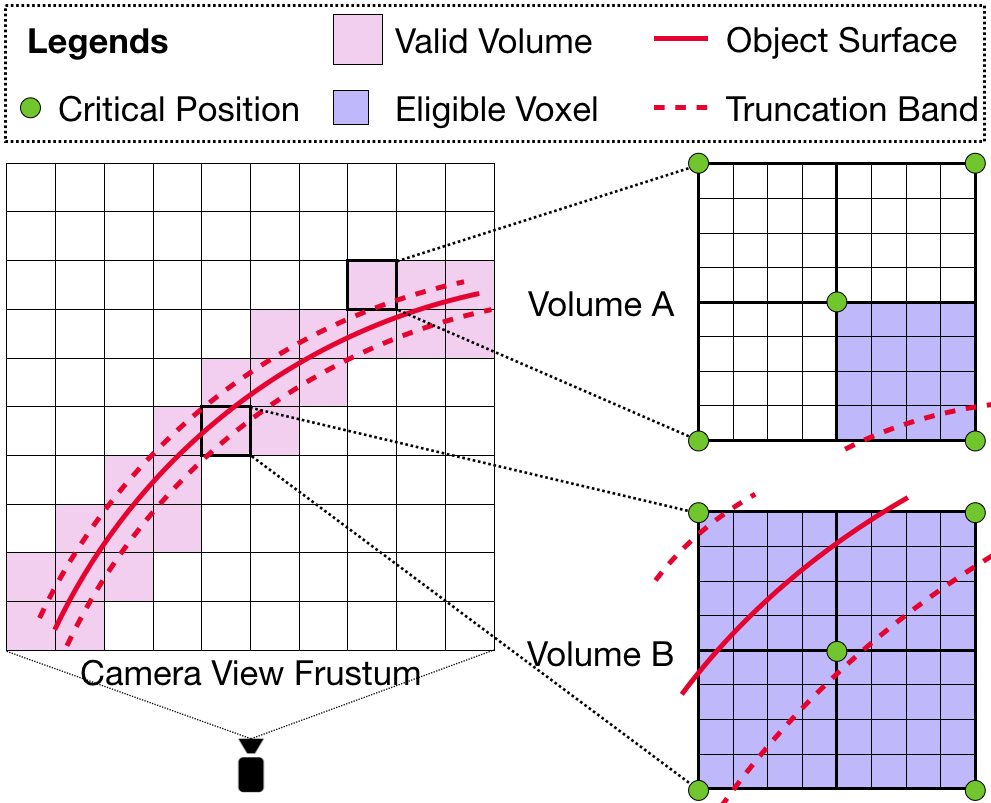}
    \caption{Our design decides eligible \textbf{voxels} (in purple) from valid \textbf{volumes} (in pink) by testing 9 critical positions (5 in 2D). A center point outside the truncation band finds 1 or 2 sub-volumes (purple in Volume A), while a center point within the band finds all voxels eligible for fusion (Volume B).}
    \label{fig:algo}
  \end{minipage}
\end{figure}

\fixme{\cref{fig:fusion-char}} shows a breakdown of the processing status of voxels. Surprisingly, only around \fixme{40\%} of voxels proceed to the TSDF calculation, while around \fixme{30\%} terminate early due to being outside the truncation band. Although processing early-terminated voxels involves minimal condition checks, the overheads before these checks (e.g., parallelizing voxels) are similar to those for fused voxels.

\textbf{Method.}
Targeting the primary inefficiency, we employ a voxel pruning design, similar to but more aggressive than FlashFusion~\cite{han2018flashfusion}, to filter out voxels outside the truncation band, based on two key insights: (1) object surfaces intersecting with a voxel are nearly flat at the voxel’s granularity (5 to 10 mm),  and (2) voxels are densely aggregated into volumes to exploit locality. It is possible to estimate voxel status inside a volume by checking only a few critical positions. Intuitively, a volume with eight corners beyond the truncation band should have zero voxels eligible for fusion.

\fixme{\cref{fig:algo}} presents a 2D cross-section of an example surface (\fixme{in red}) and nearby valid volumes (\fixme{in pink}) in a 3D scenario. To prevent cases where the entire volume is considered eligible when only a corner falls within the truncation band (Volume A), we further partition the volume into 8 sub-volumes (4 in the 2D example) by evenly dividing the cube in three dimensions. We check the center point, in addition to the eight corners of the volume (4 in the 2D example), to determine sub-volume's eligibility. A center point within the truncation band makes all voxels in the volume eligible for TSDF Fusion (Volume B), while only a subset of voxels in a volume are eligible if only a corner is within the truncation band.

Voxel pruning reduces the number of processed voxels by \fixme{50\%}, from \fixme{1.77 billion} to \fixme{0.89 billion}. \hl{We compare our design with FlashFusion{~\cite{han2018flashfusion}} by integrating their pruning logic into our codebase and measuring the total time for TSDF Fusion on the Jetson board.} Note that FlashFusion aims to find valid volumes, represented by the \fixme{pink} regions in \fixme{\cref{fig:algo}}, whereas our design improves within each valid volume. To ensure a fair comparison, we apply FlashFusion's 8-corner check to valid volumes, denoted as FF$^{\prime}$, and compare it with the baseline and our design in \fixme{\cref{fig:fusion-char}}. Our design exploits fine-grained pruning and outperforms FF$^{\prime}$ by \fixme{29.4\%}. 

\subsection{Operation Pruning}
For each eligible voxel, Open3D performs the weighted average TSDF computation as described in~\cref{sec:background}. However, the intermediate weighted average values are not required; only the final TSDF value is consumed by the downstream computation (e.g., occlusion checking or marching cubes). This observation offers the opportunity to reduce computation waste by pruning some operations, as noted in~\cite{han2018flashfusion}.
Specifically, for the $T_n$ computation in equation~\ref{eq:tsdf}, instead of performing a  division to compute an intermediate weighted average at each voxel, we keep a running sum of the weighted depth values and a sum of the weights. We then perform the division just once at the end to compute the final weighted average, thus saving several costly division operations.

\begin{figure}[t!]
  \begin{minipage}[t]{0.48\columnwidth}
    \centering
    \includegraphics[width=\columnwidth]{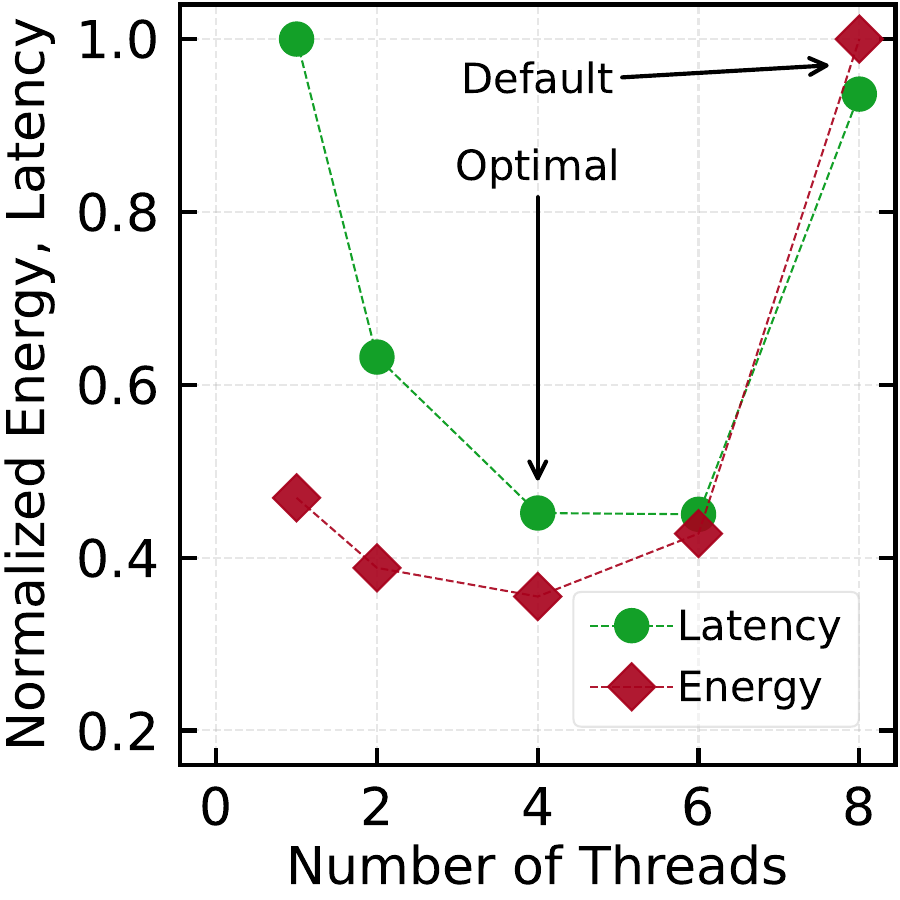}
    \caption{Energy and latency as functions of the number of threads. The sweet spot (4) in the energy-latency trade-off differs from the default (8), indicating that parallelization must be carefully implemented to balance overheads and benefits.}
    \label{fig:omp}
  \end{minipage}
  \hfill
  \begin{minipage}[t]{0.48\columnwidth}
    \centering
    \includegraphics[width=\columnwidth]{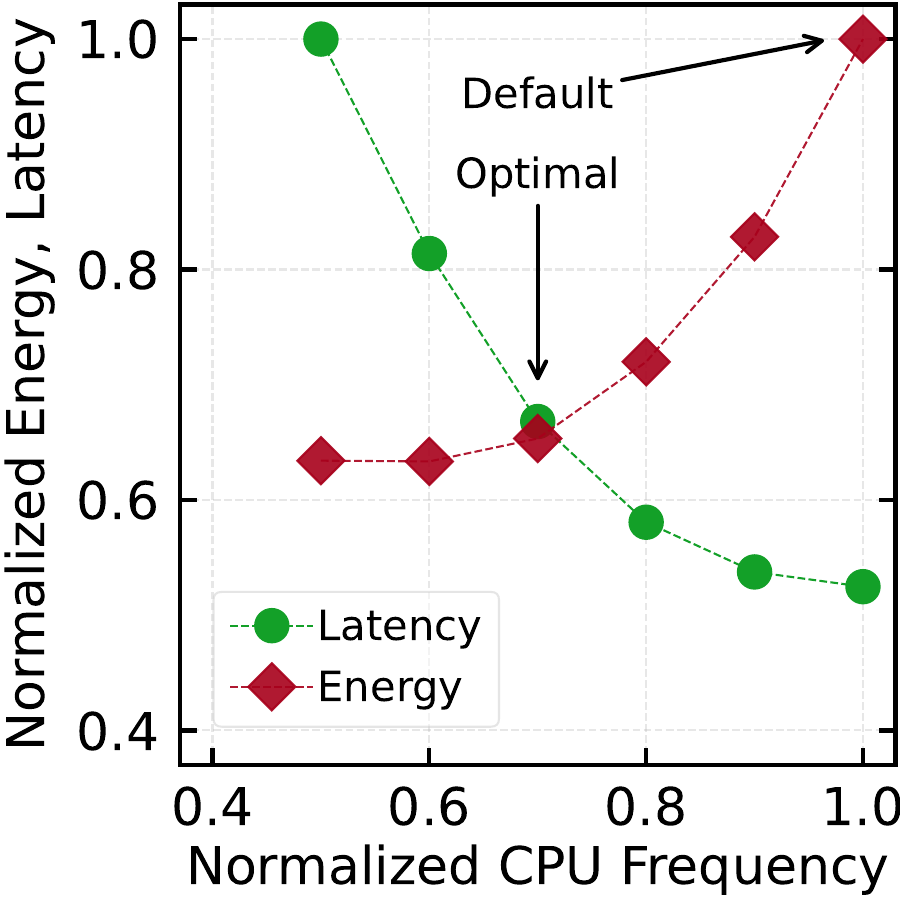}
    \caption{Energy and latency as functions of CPU frequency. Energy and latency change in opposite directions and at different rates as frequency changes, motivating careful design for an energy efficient sweet spot in the energy-latency trade-off.}
    \label{fig:cpu-freq}
  \end{minipage}
\end{figure}

\subsection{Parallelism} Many machine perception tasks employ parallelism to reduce latency. However, parallelism also incurs overheads; e.g., for thread creation, synchronization, communication, and due to load imbalance. Depending on how the algorithm is parallelized (and hardware attributes), the positive benefits of parallel execution may or may not outweigh the negative impact of parallelism overheads. Further, these impacts may affect latency differently from energy resulting in a disproportionate energy-latency trade-off. 

Our baseline Open3D implementation, configured by default to use the maximum number of processing cores, exhibits all the aforementioned impacts. As a motivational example, \cref{fig:omp} shows measured results for a ScanNet sequence running on the Nvidia Jetson Xavier board with 8 cores (more details in~\cref{sec:eval}). The figure shows measured average latency per frame and total CPU energy for the entire sequence with 1, 2, 4, 6, and 8 threads (shown as a fraction of latency and energy with 1 thread). The figure clearly shows that the default value of 8 threads is neither latency nor energy optimal. Running with 6 threads is latency optimal, but not energy efficient -- the reduction in latency from 4 to 6 threads is minor compared to the increase in energy. In this case, 4 threads provides the sweet spot to balance energy and latency. \fixme{Further analysis shows that the main reason for the low parallelism effectiveness at the default configuration of 8 threads is the relatively low amount of total work in the parallel portion, which is not able to amortize the parallelism overhead at 8 threads.}
\section{Execution Driven Energy Optimization}
\label{sec:exe}

Modern hardware provides multiple configurations to enable different energy vs.\ latency trade-offs. Processor frequency is one of the most widely used configuration parameters. The highest supported processor frequency is the optimal configuration if energy is not a consideration or if the lowest possible latency is a desirable target. 
For mobile \hl{machine} perception, however, neither is true. Energy is a first-order constraint, and we do not need to run any faster than necessary depending on downstream applications, such as human perception in AR or control loop frequency in robotics. Hence, in mobile \hl{machine} perception, {\bf faster is not always better.} 

\cref{fig:cpu-freq} shows the trade-off between energy and latency for TSDF Fusion on the Nvidia Jetson Xavier board. The figure plots CPU energy and latency for frequency spanning from half the maximum supported frequency to the maximum. As frequency increases, latency decreases but energy increases. However, the latency decrease is not proportional to energy use. At higher frequencies, a small latency benefit results in a disproportionately large energy increase. For instance, at 80\% of the maximum frequency, latency reduces by \fixme{5.6\%}, but energy increases by \fixme{28.0\%}.

This phenomenon arises due to properties of CMOS logic:
$Power \propto Voltage^2 ~ \times ~ Frequency$. Furthermore, higher frequency requires higher voltage, making the relationship between power and frequency super-linear.\footnote{The voltage-frequency relationship was linear until recent CMOS generations, resulting in a cubic power-frequency relationship. Recent CMOS generations have seen sub-linear voltage vs.\ frequency, but the super-linear power vs.\ frequency trade-off remains for significant regimes. For simplicity, we have also ignored static power in the above discussion.} Latency on the other hand reduces linearly with increasing frequency for CPU dominant workloads (sub-linear for others). Since $Energy = Power \times Latency$, latency and energy do not improve proportionally with frequency, providing an opportunity to optimize algorithms for better energy-latency trade-offs. 

Though well understood in hardware and systems communities, this energy-latency trade-off is not fully exploited by modern machine perception tasks. Effective use of this technique requires {\em slack} in the computation; i.e., if the algorithm choice is such that the frame latency is barely within the required constraint, then reducing frequency is not an option since it will entail exceeding the acceptable frame latency. The algorithm-driven techniques from~\cref{sec:algorithm} can enable precisely such a slack by reducing per frame latency, increasing the scope of benefits from the execution-driven techniques and driving the system more towards energy efficiency.

\begin{figure}[h]
    \centering
    \vspace{-5pt}
    \captionsetup[subfigure]{width=0.5\columnwidth}
    \subfloat[\small{Average F-score of all sequences.}]
    {
        \includegraphics[width=.48\columnwidth]{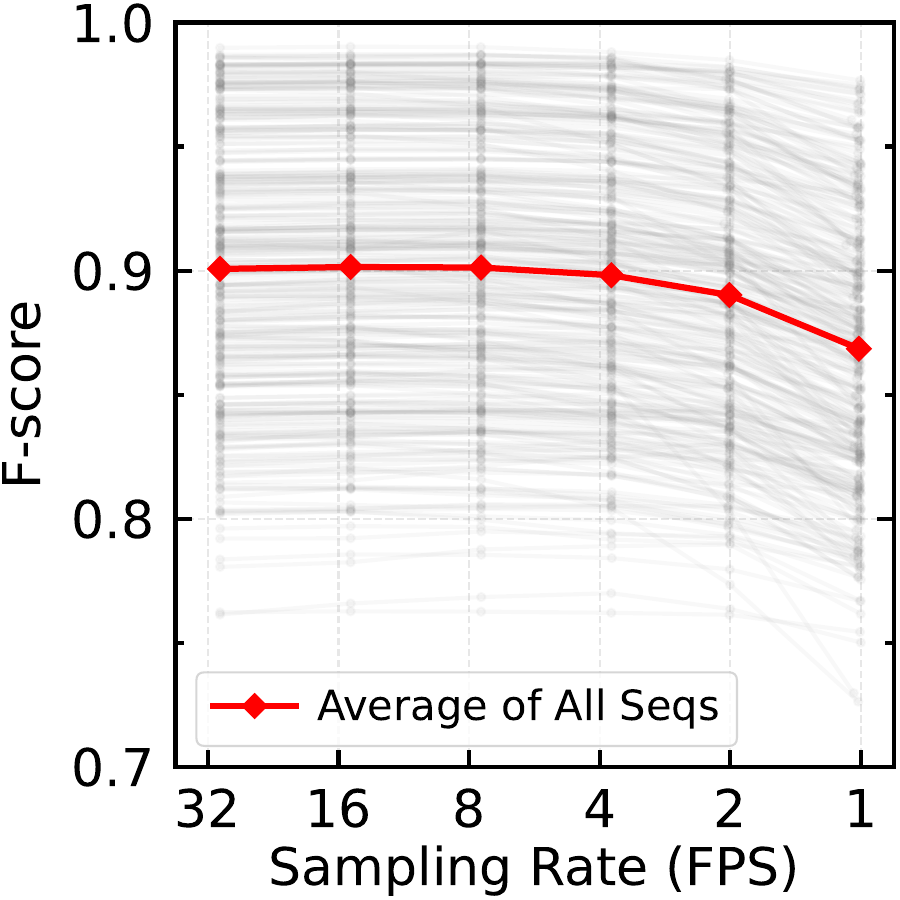}
        \label{fig:quality}
    }
    \subfloat[\small{Sequences used for evaluation.}]
    {
        \includegraphics[width=.48\columnwidth]{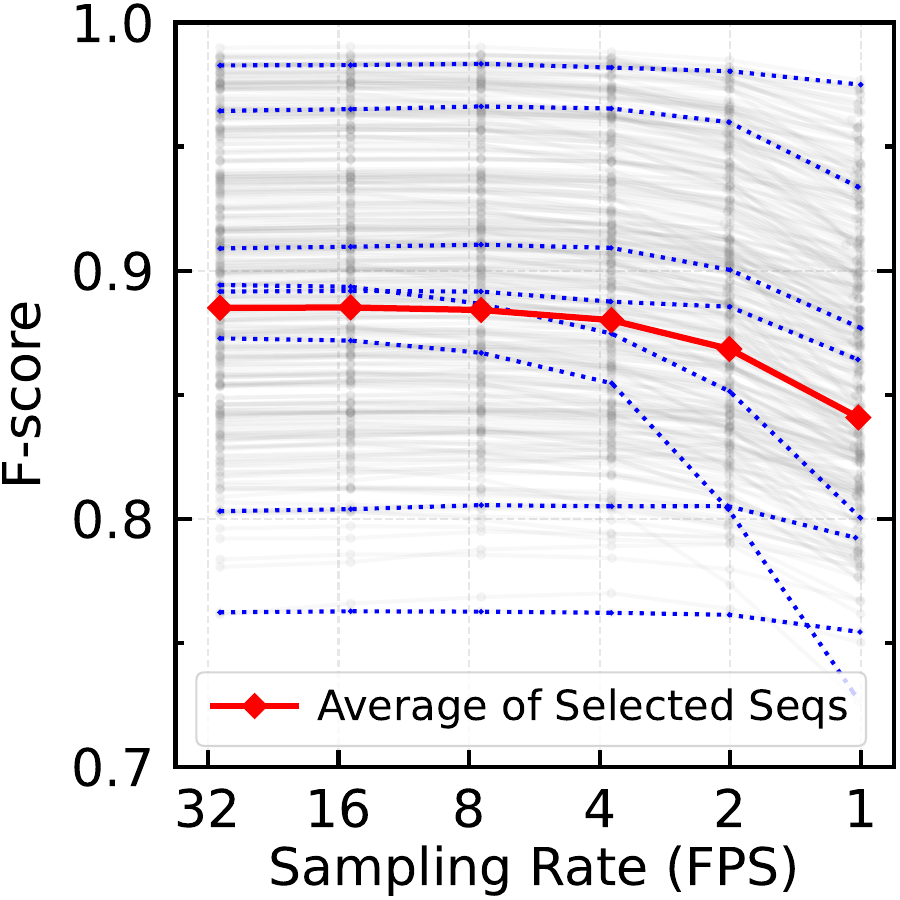}
        \label{fig:selected}
    }
    \caption{The mesh quality varies with frame sampling rate, showing minor accuracy loss in the average F-score across the ScanNet validation set up to 3.75~FPS. We choose \fixme{\hl{8}} representative sequences with averaged trends that align with the overall validation set, covering distinct trade-offs between mesh quality and sampling rate.}
    \label{fig:quality-sampling-rate}
    \vspace{-5pt}
\end{figure}
\section{Data Driven Energy Optimization}
\label{sec:data}

Sensory data across frames often exhibits significant redundancy, resulting in inefficient computations that consume energy without proportional benefits to reconstruction. Data sampling reduces redundancy by selecting frames to process, which can boost energy efficiency but may compromise the algorithm's accuracy.

Surprisingly, uniformly sampling from ScanNet's 30 frames per second (FPS) data sequence yields significant energy reduction with only a minor accuracy compromise, we therefore adopt uniform sampling to balance accuracy and energy.

The uniform sampling rate is a configurable parameter that controls the level of redundancy. Intuitively, a higher sampling rate is likely to reduce redundancy but may lead to more accuracy compromise. We quantify redundancy as the ratio of depth measurements observed in both the current and the last fused frame relative to the total valid depth measurements in the current frame. We find the original ScanNet validation set has an average redundancy of \fixme{0.96}. 

We choose 30, 15, 7.5, 3.75, 2, and 1 FPS as candidate rates for uniform sampling, corresponding to sampling every 1, 2, 4, 8, 15, and 30 frames. These candidates progressively reduce the redundancy rate from \fixme{0.96} to \fixme{0.60} and improve energy efficiency. For instance, scene reconstruction with a data stream sampled at 3.75~FPS yields meshes virtually identical to the 30~FPS baseline, while achieving significant energy reduction.

We observe a consistent trend when applying uniform sampling across all of the 312 sequences in the ScanNet validation set. \cref{fig:quality} shows F-score as a function of uniform sampling rate for each sequence (shown in grey), with the average F-score represented by the red curve. On average, accuracy decreases only slightly across all sequences until reaching 3.75 FPS, supporting the use of uniform sampling to balance accuracy and energy.
\section{Experiments}
\label{sec:eval}

\subsection{Experimental Setup}
\label{sec:eval:setup}
    We evaluate system designs incorporating a comprehensive set of combinations of the algorithm-, execution-, and data-driven energy optimizations from \cref{sec:algorithm} to \cref{sec:data} using the methodology below.

    \textbf{Hardware.}
    We use an NVIDIA Jetson AGX Xavier~\cite{Jetson_xavier_agx} development board as our evaluation platform, which features an 8-core Arm-v8 64-bit CPU, a 512-core Volta GPU, and 16GB of LPDDR4x DRAM. The power consumption of the board is in a similar range to the chipsets used in recent XR headsets~\cite{Snapdragon, Quest3Snapdragon}.

    \textbf{Datasets.}
    We use ScanNet~\cite{dai2017scannet} for characterization and evaluation. It has 1201 training, 312 validation, and 100 test sequences, each featured with high-quality ground-truth meshes, poses, and RGBD video streams recorded at 30 FPS in real-world indoor environments. We use sequences from the validation set because our experiments do not involve training or testing.

    \textbf{Sequences}
    We evaluate \fixme{\hl{8}} representative sequences from the ScanNet validation set. These sequences span a broad range of quality trade-offs, and their average trend closely aligns with that of the entire validation set (\cref{fig:selected}). \hl{The sequences cover various scene types, scales, motion speeds, complexity levels, lighting conditions, and sequence lengths, as elaborated in{~\cref{tab:eval_seq}}.}

    \begin{table*}[h!]
    \caption{\hl{Characteristics of selected evaluation sequences. We abbreviate Moderate as \textquotesingle Mod\textquotesingle ~and Medium as \textquotesingle Med\textquotesingle.}}
    \vspace{-5pt}
    \centering
    \begin{tabular}{ ccccccccc }
    \toprule[0.15em]
    Sequence Number     & \textbf{0088\_02}         & \textbf{0131\_01}     & \textbf{0378\_00}     & \textbf{0461\_00}
                        & \textbf{0474\_04}         & \textbf{0568\_00}     & \textbf{0607\_00}     & \textbf{0664\_00}
    \\ \midrule[0.15em]
    Sequence Length     & 2210          & 1122      & 1892      & 520       & 848       & 1651      & 992       & 1439
    \\ \midrule[0.05em]
    Scene Complexity    & High          & Mod-Low   & High      & Low       & Moderate  & Mod-Low   & Mod-High  & Low
    \\ \midrule[0.05em]
    Scene Type          & Conf. Room    & Office    & Office    & Lobby     & Office    & Lounge    & Kitchen   & Bathroom
    \\ \midrule[0.05em]
    Scene Scale         & Large         & Medium    & Med-Small & Med-Small & Med-Large & Large     & Medium    & Small
    \\ \midrule[0.05em]
    Motion Speed        & Moderate      & Mod-Fast  & Mod-Slow  & Moderate  & Fast      & Mod-Slow  & Mod-Fast  & Slow
    \\ \midrule[0.05em]
    Light. Condition   & Mod-Poor      & Good      & Mod-Poor  & Poor      & Poor      & Good      & Mod-Poor  & Good
    \\ \bottomrule[0.15em]
    \end{tabular}
    \label{tab:eval_seq}
    \vspace{-6pt}
    \end{table*}

    \textbf{Metrics.}
    \cref{sec:background} describes metrics we evaluated.
    
    \textbf{Baseline and Design Variants.}
    The baseline is the out-of-the-box implementation of Open3D operating at a 30~Hz frame rate, which is used for capturing the dataset. Our evaluated design variants perform a comprehensive sweep of various optimization combinations from \cref{sec:algorithm} to \cref{sec:data}. To highlight the optimizations within each design, we label them using a combination of the following notations for each constituent optimization:.
    \begin{itemize}[noitemsep, nolistsep]
        \item \textbf{A}: All {\bf A}lgorithm optimizations in \cref{sec:algorithm}.
        \item {\textbf E(X)}: {\bf E}xecution optimization of \cref{sec:exe} with processor frequency set to $X\%$ of maximum. $X \in [100, 90, 80, 70, 60, 50]$.
        \item {\textbf D(X)}: {\bf D}ata optimization of \cref{sec:data} with uniform sampling resulting in $X$ FPS. $X \in[1, ~2, ~3.75, ~7.5, ~15, ~30]$.
    \end{itemize}

    \textbf{Energy Consumption.} We apply a custom tool, similar to~\cite{ihwoo_2020}, which monitors various power channels provided by the Jetson board, and precisely records the start and end timestamps of TSDF Fusion for each frame. We average the power readings across the entire processing period, and derive energy consumption by multiplying power by the total processing time.
    
    The Jetson board internally carries two INA3221 power monitors~\cite{Jetson_agx_guide}, reporting the instantaneous voltage and current readings for the CPU, GPU, DDR, SoC, Sys, and CV components of the board. We choose to report the power consumption of workload-related components, specifically DRAM (DDR) and CPU.
    
    We exclude power data from components such as on-chip microcontrollers (SoC) and I/O (Sys). A considerable part of their power usage is due to connectivity to peripherals like cameras, displays, WiFi, etc, which are not in use in our experiments. In fact, the SoC and Sys power remains relatively constant regardless of the workload being executed or a particular design being evaluated. 

    \textbf{Experimental Details.}
    Errors in pose estimation can affect scene reconstruction accuracy, which are orthogonal to our work on TSDF Fusion. We therefore modify the Open3D codebase to use the ground-truth poses provided by the dataset. We record timestamps at the start and end of the TSDF Fusion to measure processing time and energy consumption, with disk I/O time excluded.

    We configure the processors to operate at the specified frequency with Dynamic Voltage and Frequency Scaling (DVFS) disabled to avoid unpredictable scaling overhead. We set the cooling fan speed and DRAM frequency to their maximum levels to eliminate potential memory and thermal impacts.

\begin{figure}[ht!]
    \centering
    \includegraphics[width=\columnwidth]{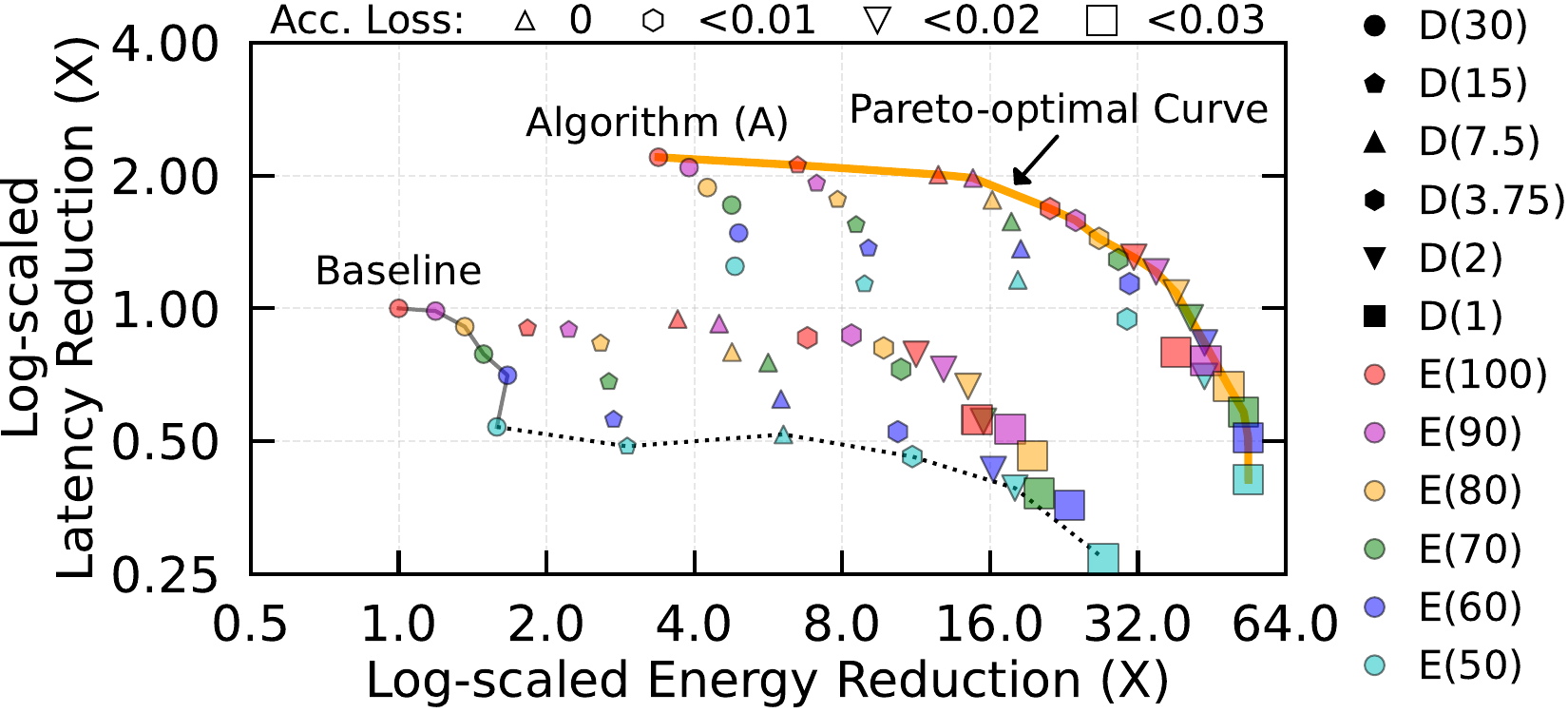}
    \caption{The trade-off space shows various parameter configurations, with each point representing a design combining $A$, $E(X)$, and $D(X)$. Colors indicate execution frequencies, shapes denote data sampling rates, \hl{and marker sizes represent the accuracy losses compared to the baseline.} The Pareto-optimal curve, $D(30)$, and $E(50)$ are connected for reference. Designs with algorithm augmentation form the upper cluster, offering a broad range of options.}
    \label{fig:dse-full}
    \vspace{-5pt}
\end{figure}

\subsection{Design Space Analysis and Implications for Energy-Efficient Design}

A three-dimensional trade-off space exists among energy, latency, and accuracy, where each design represents a point in this space with various optimizations enabled. We present and characterize this trade-off space on a mobile platform similar in form factor to commercial XR systems~\cite{VisionProBatteryCapacity}, and discuss its implications for energy-efficient designs. We evaluate 72 designs selecting between the baseline and optimized algorithm, a choice among 6 processor frequencies ([100, 90, 80, 70, 60, 50] percent of the maximum), and 6 data sampling rates ([30, 15, 7.5, 3.75, 2, and 1] FPS). Scaling experiments on a mobile platform is challenging, especially for intensive workloads like scene reconstruction. Therefore, we examine all 72 designs on a sample data sequence (0153\_00) and perform a sensitivity study for 8 representative sequences in \cref{sec:eval:sens}.

\textbf{\textcircled{\small{1}}A Broad Trade-off Space.}
Overlooking the broad trade-off space often results in sub-optimal designs that fail to achieve energy efficiency. \cref{fig:dse-full} visualizes this trade-off space by projecting designs onto the energy-latency plane and representing accuracy loss with marker sizes. We show the reduction rates for energy and latency, as well as the absolute differences in F-score for accuracy, all compared against the baseline.

In addition to size, each design is distinguished by color and shape. Designs with the same color were executed at the same processor frequency, while those with the same shape were executed at the same data sampling rate. This relationship is exemplified by two connected lines in~\cref{fig:dse-full}: a solid gray line for $D(30)$ and a dotted black line for $E(50)$. The Pareto-optimal curve for latency and energy, annotated in orange, highlights designs that offer superior reductions in latency or energy compared to those below it.

Our evaluated designs show a more varied distribution for latency and energy compared to accuracy, with latency changes ranging from a 4$\times$ slowdown to a 2$\times$ speedup, energy savings of up to 60$\times$, and accuracy loss of up to 0.03. The widely distributed design points exhibit distinct priorities. For instance, designs on the left side of the curve, such as $A+E(100)+D(30)$, prioritize minimizing latency while neglecting a potential order of magnitude in energy reduction. Conversely, designs on the right side, like $A+E(60)+D(1)$, achieve substantial energy reductions but come with a notable increase in latency. In fact, the common practice of executing programs at the maximum processor frequency without data sampling aligns with the $A+E(100)+D(30)$ design mentioned above. The flat Pareto-optimal curve near this design indicates that the potential to reduce energy usage with negligible latency increase is being missed.

\textbf{\textcircled{\small{2}} Importance of Co-optimization.} 
Achieving a design on the Pareto-optimal curve requires co-optimizing processor frequency and data sampling rate. Algorithm optimization is always beneficial since it does not compromise latency or accuracy. The Pareto-optimal curve is jointly formed by designs with higher processor frequencies and lower data sampling rates. Relying on a single optimization alone is inadequate for accessing the full range of the curve and may lead to sub-optimal designs.

\begin{figure}[ht!]
    \centering
    \includegraphics[width=\columnwidth]{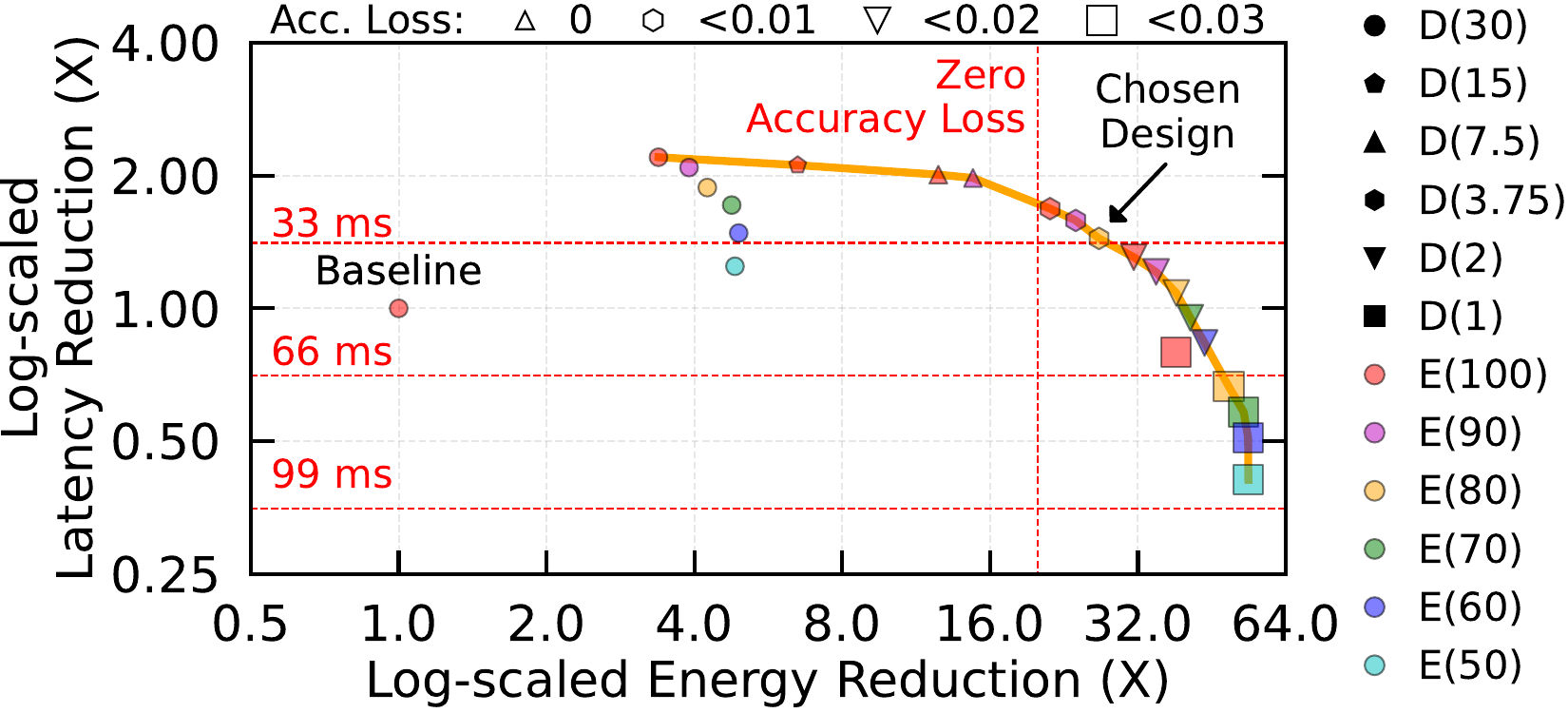}
    \caption{The simplified trade-off space. The Pareto-optimal curve can only be fully accessed through co-optimization. Constraints on latency and accuracy directly impact the accessible trade-off space. Common practice, which uses a latency constraint of 33~ms (30~FPS) and 0 accuracy loss, limits the trade-off space to the top-left area, missing much of the Pareto-optimal curve. Releasing accuracy constraints allows access to the top-right corner. Finding optimal constraint values requires cross-component analysis. We evaluate the annotated design in \cref{sec:eval:sens} for sensitivity analysis.}
    \label{fig:dse-annotate}
\end{figure}

To illustrate further, \cref{fig:dse-annotate} shows a simplified trade-off space with designs on the Pareto-optimal curve, alongside designs optimized in isolation for processor frequency or data sampling. Within the $D(30)$ cluster, $A+E(60)$ is the lowest energy design without considering data sampling. However, a set of designs in the top right direction, which includes data sampling, achieves additional energy reduction ranging from 1.32$\times$~($A+E(100)+D(15)$) to 5.42$\times$~($A+E(80)+D(3.75)$). Similarly, $A+D(1)$ is the lowest-energy design without frequency control but has \fixme{1.14$\times$} higher energy and \fixme{1.05$\times$} higher latency compared to $A+E(60)+D(2)$, which combines processor frequency tuning for better efficiency.

The need for co-optimization arises from the distinct latency-energy trade-offs of each individual optimization. In the left range of the Pareto-optimal curve, reducing the data sampling rate with $D(X)$ yields significant energy savings with minor latency penalties. In contrast, applying $E(X)$ in the right range of the curve shows an opposite trend. This transition shows how the two optimizations collaboratively shape the design frontier, calling the need for co-optimization to fully explore the available design candidates.

\textbf{\textcircled{\small{3}} Tailored Constraints from Cross-component Analysis.} Design candidates within the trade-off space are constrained by latency and accuracy requirements. Ill-defined constraints may unnecessarily limit the potential candidates, thereby leading to sub-optimal results. Achieving optimal designs necessitates cross-component analysis to align constraints with the needs of downstream consumers, such as algorithms, components, or end users.

Cross-component knowledge enables additional energy benefits that might be overlooked. We discuss the impact on accuracy and latency constraints, respectively. Reference lines are added to the trade-off space to indicate latency constraints at frame intervals of 30~FPS (camera frame rate, equivalent to 33~ms), 15~FPS (66~ms), and 7.5~FPS (99~ms). We also show the boundary between designs with and without accuracy loss.

\underline{Accuracy Constraints.} Without cross-component analysis to guide accuracy constraints, pursuing zero accuracy loss may exclude candidates with data sampling rates lower than $D(7.5)$. While $A+E(70)+D(7.5)$ is the lowest energy option under the initial constraints (i.e., 33~ms camera frame rate and zero accuracy loss) and achieves \fixme{$\sim16\times$} energy reduction, additional savings from trading accuracy are missed, ranging from about \fixme{20$\times$ to 60$\times$}. For instance, $D(3.75)$ brings an extra \fixme{2$\times$} energy savings with a negligible F-score accuracy loss of \fixme{less than 0.01}.

\underline{Latency Constraints.} Similarly, without analyzing and relaxing \textbf{latency} constraints from consumer components, even a relaxed \textbf{accuracy} constraint only enables three additional design candidates, namely $A+E(80\sim100)+D(3.75)$. Latency constraints are initially imposed as the time interval between captured frames, preventing future frames from being blocked and ensuring prompt content updates.
However, with a sampled input data stream, frames are unlikely to be blocked, as even $D(15)$ increases the frame interval to $\sim$66~ms. Therefore, delay-tolerant consumers could further relax latency constraints, allowing for designs with longer latencies. For instance, relaxing the latency constraint to 66~ms shifts the design choice from $A+E(80)+D(3.75)$ to $A+E(60)+D(2)$, expanding energy reduction from \fixme{26.66$\times$} to \fixme{43.75$\times$}.

\underline{Summary.} Relaxing accuracy and latency constraints unlocks access to design options that bring additional energy efficiency. While precise constraints are only attainable through analysis of consumers in an end-to-end system, we exemplify two real-world use cases with divergent design constraints: one with stringent latency and accuracy requirements while the other not, to illustrate how the optimal design decision will shift.

\subsection{Application-Dependent Constraints: Case Studies}
\label{sec:eval:usecase}

\textbf{Scanning and Sharing.}
Scanning a target of interest and sharing it with friends is a common scenario but expects prompt feedback and high-quality reconstruction. High accuracy and low latency are two prerequisites that allow users to instantly know whether the target of interest is well reconstructed and rescan a compromised area if necessary. High accuracy ensures the reconstruction faithfully replicates the physical object in the virtual space, and low latency enables smooth interaction when users change their view port.

Without compromising scanning quality and processing delay from when a frame is captured, the accuracy and latency are constrained to zero loss and 33~ms, respectively, making $A+E(90)+D(7.5)$ an optimal design. Compared to the baseline, this design achieves about 16$\times$ energy reduction while \hl{maintaining accuracy and latency, thus extending the battery life of the mobile system.}

\textbf{Spatial Audio.}
Spatial audio provides an immersive acoustic experience by creating pose-dependent sound based on head position, orientation, and surrounding information. The key computation is to mimic how sound behaves in physical environments, relying on scene reconstruction to replicate the physical world in virtual space and compute acoustics like sound wave reflections and absorptions.

Scene reconstruction for spatial audio often has relatively relaxed accuracy and latency constraints, as changes in reconstruction are more noticeable in the visual domain than in the acoustic domain~\cite{kim2022immersive,SpatialAudio, schissler2017acoustic}. For instance, a misaligned wall in the reconstruction creates obvious visual artifacts but may not significantly affect sound effects. Likewise, relaxed latency constraints in scene reconstruction are unlikely to impact the acoustic experience.

Tolerance to accuracy loss and latency increase unlocks designs with $D(3.75)$, $D(2)$, and $D(1)$, allowing energy reductions ranging from $\sim20\times$ to over 60$\times$. As the time interval between frames at $D(3.75)$ ($\sim$264ms) is already far longer than the latency of any candidates, frames are guaranteed to be finished before the next frame arrives. Therefore, the optimal design can be $A+E(60)+D(1)$ with \fixme{$\sim$0.03} accuracy loss and \fixme{$\sim$60$\times$} energy reduction. The effect of relaxing latency constraints is significant, as otherwise only three candidates, $A+D(3.75)$ with $E$ ranging from 100 to 80, are viable.

{\bf Intermediate Case.} For more diversity of use cases, we introduce a third scenario with constraints between those of scanning/sharing and spatial audio.
The latency target is restricted to the baseline sensor frame interval (33~ms), and the accuracy target tolerates a 0.01 drop in F-score. These constraints narrow the trade-off space, leading us to $A+E(80)+D(3.75)$ as the optimal design (denoted in \cref{fig:dse-annotate}). This design observes a notable energy reduction of 26.66$\times$, a 1.44$\times$ decrease in frame latency, and a negligible accuracy loss of 0.002 in F-score, relative to the baseline.

\underline{Summary.} The optimal design is application-dependent, as different constraints on latency and accuracy lead to divergent design choices. Applying one design across various applications may either violate the constraints expected by downstream consumers or miss energy reduction opportunities, as evidenced by the optimal designs identified for the above three use cases.

\subsection{Sensitivity Analysis}
\label{sec:eval:sens}
The large design space limited our previous analysis to a single data sequence. We now expand evaluation to 8 representative sequences with distinct characteristics, as detailed in~\cref{sec:eval:setup}. We focus on the intermediate (third) use case and its identified optimal design.

On average across 8 evaluated sequences, with only a minimal compromise in mesh quality (\fixme{0.005} in F-score), the chosen design achieves a notable energy reduction of \fixme{24.90}$\times$ and a decrease in frame latency of \fixme{1.45}$\times$. These improvements are attainable only through co-optimization of processor frequency and data sampling rate, as evidenced by the breakdown of each optimization below.

\begin{figure}[ht]
    \centering
    \captionsetup[subfigure]{width=\columnwidth}
    \subfloat[\small{Energy reduction over the baseline.}]
    {
        \includegraphics[width=0.98\columnwidth]{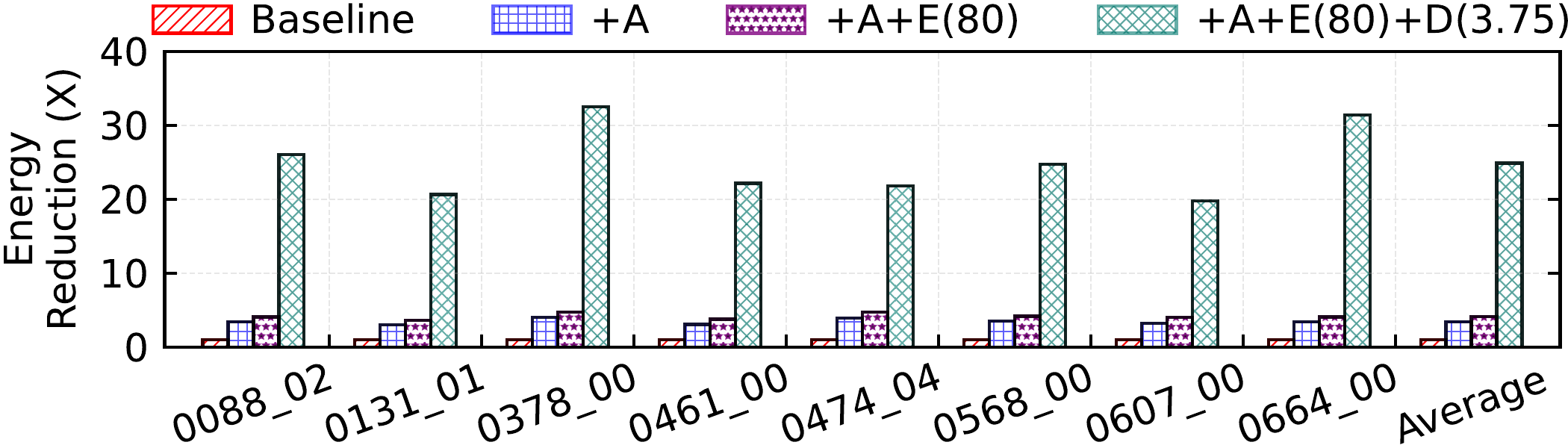}
        \label{fig:main-energy}
    }
    \\
    \captionsetup[subfigure]{width=\columnwidth}
    \subfloat[\small{Per-frame latency over the baseline.}]
    {
        \includegraphics[width=0.98\columnwidth]{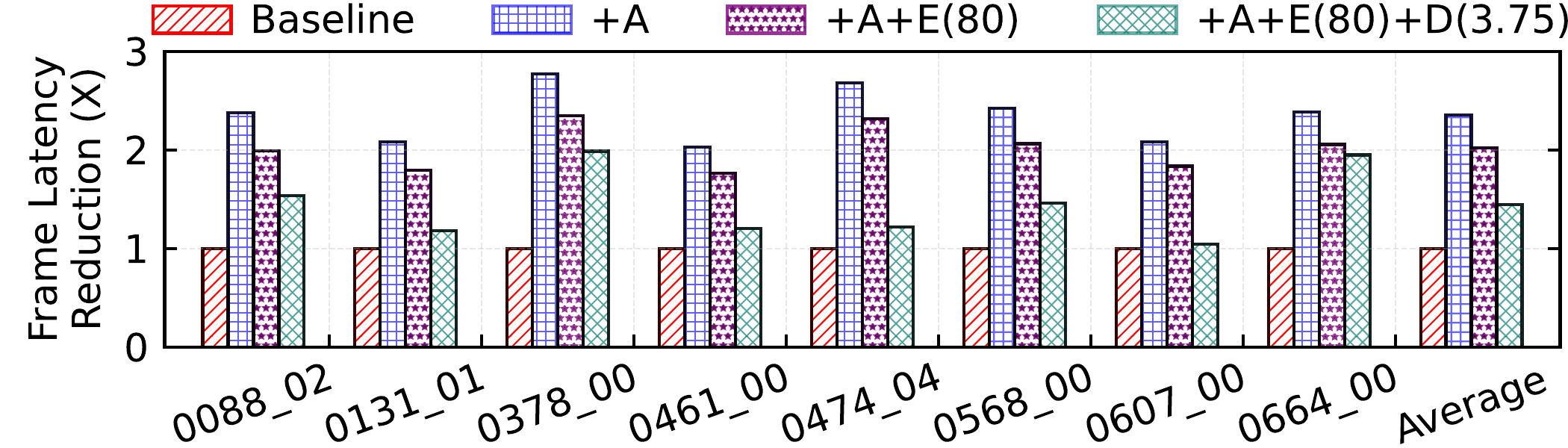}
        \label{fig:main-latency}
    }
    \caption{Energy reduction increases progressively with optimizations applied. $D$ achieves the greatest reduction at the cost of accuracy and frame latency, while $A$ effectively compensates for the latency penalty. Consistent trends across sequences emphasize the importance of multifaceted co-optimization.}
    \label{fig:main-result}
    \vspace{-5pt}
\end{figure}

\textbf{Energy.} Energy reduction increases progressively as optimizations are enabled (\cref{fig:main-energy}). The chosen design achieves a \fixme{24.90}$\times$ energy reduction, with contributions of \fixme{3.45}$\times$ from $A$, \fixme{1.21}$\times$ from $E(80)$, and \fixme{5.98}$\times$ from $D(3.75)$. While mitigating data redundancy drives most of the benefits, the other two optimizations are also crucial, as every Joule is valuable in battery-constrained XR systems.

\textbf{Latency.} The reduction in frame latency clearly illustrates the latency-energy trade-off and the need for co-optimization (\cref{fig:main-latency}). While both $E$ and $D$ sacrifice latency for energy savings, these penalties are compensated by $A$, which still leads to an average \fixme{1.45}$\times$ latency reduction. $E$ trades latency for energy by scaling down processor frequency, boosting energy reduction from \fixme{3.45}$\times$ to \fixme{4.16}$\times$ but reducing frame latency improvement from \fixme{2.35}$\times$ to \fixme{2.02}$\times$. On the other hand, $D$ prolongs frame latency because partial workloads from skipped data cannot be eliminated but are merely deferred to the next processed frame. For instance, a voxel that should have been allocated earlier must now be created when processing the sampled frame to enable TSDF Fusion.

\hl{\textbf{Ablation.} The three algorithmic enhancements are evaluated collectively to avoid overemphasizing the importance of specific designs while focusing on demonstrating the trade-off space. We further break down their contributions and show that each of them enables a considerable portion of the benefits. Specifically, the latency and energy reductions unlocked by voxel pruning, operation pruning, and mitigating parallelism overhead are [30.56\%, 36.60\%, 32.84\%] and [27.48\%, 26.12\%, 46.40\%], respectively.}

Among all tested sequences, the chosen design consistently demonstrates significant energy and latency reductions, highlighting the effectiveness and generalizability of selecting the right design point from the wide trade-off space. Beyond the scene reconstruction we studied in this paper, similar trade-off spaces and optimizing opportunities in $A$, $E$, and $D$ are prevalent in many other applications. For instance, object detection and panoramic segmentation are critical \hl{machine} perception tasks in XR systems. Operation redundancy and parallelization overheads ($A$) can appear as less significant branches in neural-based algorithms. $E$ remains algorithm-independent, and data redundancy in $D$ is usually present when data comes from continuously operating sensors.

\subsection{Subjective Evaluation}
While we use the widely adopted F-score as a quantitative metric to evaluate the accuracy (i.e., mesh quality) of scene reconstruction, we also assess the subjective human-perceived quality of the chosen design by presenting side-by-side comparisons of the meshes.

Due to space constraints, \fixme{{\cref{fig:quality-evaluation}}} shows four scenes from the ground-truth and the baseline algorithm operating at a 30~FPS dataset frame rate. The reconstructions of $A+E(80)+D(3.75)$ exhibit high visual similarity to these references, with a clear and cohesive appearance of objects. Although the meshes display slight uneven boundaries, particularly at the top of the wall, this is not typically a critical area for applications consuming the mesh. Zoomed-in sections highlight subtle differences. Despite being slightly blurred compared to the ground-truth, the meshes from the chosen design closely resemble those from the 30~FPS baseline, which represents the highest quality achievable with the given algorithm.

The evaluated design has a conservative accuracy constraint ($<0.01$) to show how understanding the trade-off space aids in achieving energy-efficient designs. The high resemblance of reconstructed meshes in the qualitative comparison reveals a key insight: data sampling mainly affects boundary areas, which often receive less attention from a user’s perspective. This suggests that accuracy could potentially be further traded for energy savings as long as critical regions are preserved and user experience is not impacted. However, this also shows the need to understand constraints from an end-to-end system perspective.

\subsection{Limitations}
\hl{We evaluate our work on ScanNet, which, while widely adopted for scene reconstruction, has some limitations. (1) The datasets we study may not always be representative for broader XR user cases. While the high data redundancy exhibited is expected in XR, these sequences may not reflect XR user motion patterns, which contain a mix of saccadic and stationary motions~\cite{bischof2023eye,david2020effects,hu2017head}. Although we mitigate redundancy with uniform sampling, advanced sampling policies could be developed to better exploit redundancy in XR motions. (2) ScanNet's ground-truth meshes may lack sufficient accuracy, potentially biasing the design evaluation if the ground-truth is worse than both designs being compared. For instance, a 4~cm voxel ground-truth mesh cannot effectively differentiate between a 3~cm and a 1~cm mesh. Our parameter choice avoids this issue by using a 5~mm voxel when the ground-truth uses 4~mm. However, we point this out as a consideration that should be taken into account. (3) We evaluate a limited number of sequences mainly due to the time-consuming nature of averaging energy measurements from multiple rounds of scene reconstruction on mobile platforms. While results across various sequence characteristics are consistent, expanding the sequence scope could further solidify the conclusions.}
\section{Conclusion}
\label{sec:conclusion}
As XR technologies proliferate, the need for energy-efficient design becomes paramount, especially in battery-constrained mobile XR devices. This paper addresses the trilemma among the dimensions of energy, latency, and accuracy for scene reconstruction, a critical component of XR systems. We demonstrate that achieving energy efficiency requires a holistic approach that considers all three dimensions simultaneously.
We expose a broad trade-off space through multiple energy-oriented optimizations with configurable parameters. Our measurement and analysis on an embedded platform show that this space supports designs with a wide range of trade-offs, with the Pareto-optimal frontier accessible only through synergistic co-optimizations and consideration of constraints imposed by downstream consumers of scene reconstruction. Through an exploration of multiple use cases, we show energy benefits of up to 60X over the baseline, with acceptable latency and accuracy for the downstream application.
The insights gained from this study are broadly applicable to other machine perception tasks, and emphasize the importance of a comprehensive system perspective in developing energy-efficient XR systems.

\section*{Acknowledgement}
\label{sec:acl}

We thank the reviewers and Jay Cha for their valuable feedback. This work is supported in part by the IBM Illinois Discovery Accelerator Institute (IIDAI) and the National Science Foundation under grants 1956374, 2120464, 2217144, 2312102, and 2331878.

\bibliographystyle{abbrv-doi}
\bibliography{main}
\end{document}